%% file: main.tex
\documentclass[letterpaper, 10 pt, conference]{ieeeconf}  

\IEEEoverridecommandlockouts                              

\title{\LARGE \bf A Model Predictive Path Integral Method for Fast, Proactive, and Uncertainty-Aware UAV Planning in Cluttered Environments}
\author{Jacob Higgins, Nicholas Mohammad, Nicola Bezzo
\thanks{Jacob Higgins, Nicholas Mohammad and Nicola Bezzo are with the Department of Electrical and Computer Engineering, University of Virginia, Charlottesville, VA 22903, USA 
        {\tt\small \{jdh4je, nm9ur, nbezzo\}@virginia.edu}}%
}%

\newcommand{\subparagraph}{}

\usepackage{graphicx}
\usepackage{epstopdf}
\usepackage{amsmath}
\usepackage{amssymb}
\usepackage{subfigure}
\usepackage{multirow}
\usepackage{pbox}
\usepackage{algorithm}
\usepackage{algpseudocode}
\usepackage{titlesec}
\usepackage{bm}
\usepackage{url}
\usepackage{dblfloatfix} 
\usepackage{float}

\usepackage{algpseudocode,algorithm,algorithmicx}  
\algrenewcommand\algorithmicrequire{\textbf{Precondition:}}  
\algrenewcommand\algorithmicensure{\textbf{Postcondition:}}

\renewcommand{\baselinestretch}{0.935}

\usepackage{xcolor}

\DeclareMathOperator*{\argmin}{arg\,min}

\begin{document}

\graphicspath{ {./figs2/} }
\maketitle
\input{abstract}

\input{introduction}

\input{rel_lit}

\input{problem}
\input{approach}

\input{simulation}

\input{experiment}
\input{conclusion}
\input{acknowledgement}
\bibliographystyle{IEEEtran}
\bstctlcite{IEEEexample:BSTcontrol}
\bibliography{library}

\end{document}

%% file: abstract.tex

\begin{abstract}

Current motion planning approaches for autonomous mobile robots often assume that the low level controller of the system is able to track the planned motion with very high accuracy. In practice, however, tracking error can be affected by many factors, and could lead to potential collisions when the robot must traverse a cluttered environment. To address this problem, this paper proposes a novel receding-horizon motion planning approach based on Model Predictive Path Integral (MPPI) control theory -- a flexible sampling-based control technique that requires minimal assumptions on vehicle dynamics and cost functions. This flexibility is leveraged to propose a motion planning framework that also considers a data-informed risk function. Using the MPPI algorithm as a motion planner also reduces the number of samples required by the algorithm, relaxing the hardware requirements for implementation. The proposed approach is validated through trajectory generation for a quadrotor unmanned aerial vehicle (UAV), where fast motion increases trajectory tracking error and can lead to collisions with nearby obstacles. Simulations and hardware experiments demonstrate that the MPPI motion planner proactively adapts to the obstacles that the UAV must negotiate, slowing down when near obstacles and moving quickly when away from obstacles, resulting in a complete reduction of collisions while still producing lively motion.

\end{abstract}

%% file: introduction.tex

\section{Introduction} \label{sec:intro}

The interest in autonomous mobile robots (AMR) is fast growing in the private. military, and commercial sectors for its promise to revolutionize key components of many industries, such as logistics, structural inspection and transportation. For these applications, robust autonomy is the key that enables such operations to rapidly scale while keeping human intervention to a minimum, allowing such endeavors to remain affordable. The robots that are deployed in these real-world situations, however, are subject to many sources of uncertainty that introduce risks while in motion, e.g. dynamic obstacle position and disturbances. To compensate for this uncertainty, many approaches are receding-horizon in nature, so that newly-obtained information may affect the resulting trajectory. In particular, receding-horizon motion planning has recently grown into a large field of study within the robotics community.

\begin{figure}[t!]
\centering
\includegraphics[width=0.47\textwidth]{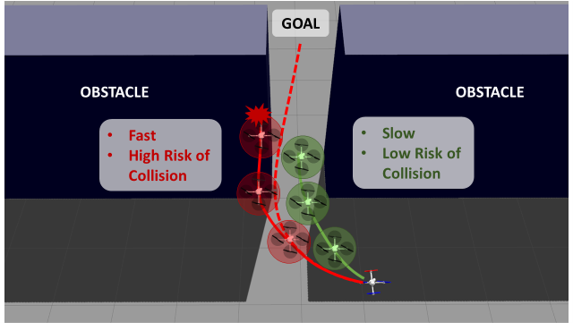}
\caption{Motivating example in which a slower speed results in safer motion through a small gap.}
\vspace{-15pt}
\label{fig:intro_pic}
\end{figure}

Typical receding-horizon approaches decouple motion planning and trajectory tracking, and are often treated as separate problems~\cite{8968021}. Such decoupling could lead to potentially dangerous situations if the motion planner is not aware of the limits and capabilities of the lower-level planner. Consider the scenario depicted in Fig.~\ref{fig:intro_pic} in which an aerial robot must pass through a narrow opening to the other side. A receding-horizon motion planner may command a  fast-moving trajectory to reduce travel time, but such a trajectory may not be perfectly tracked by the low level controller. This could induce a large tracking error, meaning that tracking a collision-free trajectory may not result in collision-free motion. In order to mitigate the risk of collision under this kind of uncertainty, the robot must command slower motion through the gap, reducing the tracking error.

One possible solution is a data-informed approach, where the tracking error and subsequent risk of collision are inferred from past performance of the robot. This data-informed risk assessment allows the risk measure to accurately reflect the performance of the low level controller, but must capture a potentially complex relationship between the commanded trajectory and the risk of tracking that trajectory. Gradient-based and quadratic-programming-based approaches are restrictive in that risk-based costs must have certain numerical qualities for real-time use. Alternatively, sampling-based approaches consider costs with minimal assumptions, allowing a greater flexibility and generality when defining risk. For this reason, the heart of the proposed approach in this paper is a receding-horizon Model Predictive Path Integral (MPPI) motion planner, adapted from the sampling-based MPPI control used in information theoretic control theory~\cite{williams2018robust}. Typical MPPI works by rapidly sampling the low level control space of the system around a ``best guess'' of the optimal open-loop control policy, and a weighted sum is performed to iteratively update this best guess, converging to the optimal control policy after many iterations. One consequence of sampling within the low level control is the need for a large number of samples at a high rate (typically on the order of 50-100~Hz). For real-time use, this requires the robot to have a GPU on board to speed up the sampling time.

Our approach adopts MPPI control to a trajectory planning setting; instead of sampling within the control space, we sample within a trajectory parameter space. In particular, the proposed MPPI trajectory planner determines waypoints that define a spline-based trajectory. This reduces the sampling space dimension, allowing our MPPI path planner to run on a CPU in real-time, hence for a more relaxed set of system requirements. Overall, our approach allows for a computationally more efficient method to sample trajectories within the MPPI framework, without sacrificing the ability to generate fast and safe trajectories.

This paper presents two main contributions. First, the MPPI control approach is cast as a parameterized high level planner, reducing the dimension of the optimization space and mitigating the hardware requirements needed to find reasonable solutions to the motion planning problem. Second, a data-informed risk measurement is included inside the cost function of the MPPI motion planner, using the actual trajectory tracking performance of the system to determine safe and lively trajectories. The result is motion that minimizes risk of collision due to trajectory tracking error while avoiding obstacles and moving towards a goal. Together, the overall approach provides a practical method for run-time risk-aware trajectory generation towards safe navigation. While this approach is flexible enough to be applied in a general motion planning setting, this paper focuses on motion planning for an unmanned aerial vehicle (UAV) since these types of systems can be greatly affected by tracking error, and such tracking error can induce risk of collision, especially when navigating cluttered environments.

The rest of the paper is organized as follows: in Sec.~\ref{sec:rel_lit} we provide an overview of the current state of the art in receding-horizion motion planning, risk-aware motion planning and sampling-based motion planning. In Sec.~\ref{sec:problem} we formally define the problem of fast and safe trajectory generation under the risk of collision due to trajectory tracking error. Sec.~\ref{sec:approach} describes the components of the proposed approach which is validated with simulations in Sec.~\ref{sec:simulation} and physical experiments in Sec.~\ref{sec:experiment}. Finally, we draw conclusions and discuss future work in Sec.~\ref{sec:conclusion}.

%% file: rel_lit.tex
\section{Related Work} \label{sec:rel_lit}

Motion planning remains an active field of research within the robotics community. For agile robotic systems such as quadrotors, most research effort is focused on leveraging this agility by creating fast, aggressive maneuvers~\cite{7299672}. Cutting-edge agile motion planning often uses gradient-based optimization~\cite{9966054}, mixed-intenger quadratic programming~\cite{8968021}, sample- and search-based methods~\cite{9721033}, or some mixture of these techniques
~\cite{9986606}.
One assumption implicit within these approaches is the ability of a low level controller to accurately track the aggressive trajectories.
Many factors, however, can contribute to a gap between the generated trajectory and the actual trajectory. For example, despite the existence of sophisticated nonlinear controller techniques~\cite{lee2010control}, linear PID low level controllers offer sub-optimal performance but remain ubiquitous in many real-world applications for low level tracking due to speed and ease of implementation~\cite{8408892}. Aerodynamic effects~\cite{9779449,9143160} and errors within the visual odometry pipeline~\cite{8461133,9863844} are additional factors that can degrade trajectory tracking, which in turn introduces a risk of collision.


One possible approach is to use data-driven models to compensate for this uncertainty, such as end-to-end neural network policies~\cite{loquercio2021learning}, self-adjusting system models~\cite{9361343}, or disturbance estimation and rejection~\cite{9624945}.
More analytical approaches seek to provide rigorous safety guarantees, such as Tube-MPC~\cite{rawlings2017model,9762493}, chance-constrained optimization techniques~\cite{9805817}, or risk-constrained motion planning approaches~\cite{8767973}. In these cases, simplifying assumptions are usually placed on the risk involved, or the policy is too computationally intensive for real-time control.

Sampling-based techniques, such as RRT-based techniques~\cite{9772994} and  motion primitive sampling ~\cite{primitives2020Kostas,9564801}, offer approximate solutions to otherwise intractable problems, especially when solutions are required in real time. One particular sampling-based technique named STOMP~\cite{5980280} computes trajectories for robotic arm manipulators by stochastic-approximated gradient descent. Such a technique allows motion planning with non-differentiable costs, but is not receding-horizon and requires knowledge of the entire configuration space to perform well, making it not suitable for AMR applications. This work eventually evolved into model predictive path integral (MPPI) control~\cite{williams2018robust,8558663,7989202}, which directly samples within the low level control space of the robot and is theoretically motivated with information-theoretic control techniques. In order to achieve aggressive maneuvers, MPPI control requires approximately 10,000 samples to be taken at a rate of 40~Hz, necessitating the use of a GPU to parallelize and speed up the computation.

Our approach leverages the ability of the MPPI control technique to optimize non-differentiable or complex costs to incorporate data-informed risk within the cost function. Specifically, we examine the risk of collision due to tracking error of a low level controller when aggressive, high-speed trajectories are planned by the MPPI path planner. In order to generalize this procedure and mitigate the number of samples required per MPPI iteration, these trajectories are parameterized over carefully-chosen polynomials, allowing the MPPI procedure to be performed in real-time on a CPU. To the best of our knowledge, our work is the first to utilize MPPI techniques to generate trajectories in a receding horizon fashion while optimizing for a risk cost that does not need to be differentiable.

%% file: problem.tex
\section{Problem Formulation} \label{sec:problem}  

In this work, we are interested in creating a receding-horizon trajectory generation policy that addresses the risk of collision due to tracking error between the commanded trajectory and the actual trajectory of a UAV, especially when navigating potentially cluttered environments. Additionally, this trajectory generation policy should be able to handle data-informed functions of risk, and consider potentially complex relationships by placing minimal assumptions on the properties such functions may have (e.g. smoothness, differentiability). We separate this problem into two parts: (i) creation of a receding-horizon trajectory generator that can optimize for a general cost function at run time, and (ii) the inclusion of a risk factor that, when minimized, commands safe and lively trajectories in the presence of low level tracking error.


\textbf{Problem 1: \textit{Receding Horizon Trajectory Generation}:}
We seek a policy $\mathcal{P}_\tau(\bm{x}(t_0))$ that takes in the current state of the robot $\bm{x}(t_0)\in\mathbb{R}^{n_x}$ and at run time returns a time-based trajectory $\tau(t)$ defined over a future horizon $t\in[t_0,t_0+t_H]$ for a low level controller to track.
This trajectory should move the robot closer to a goal state $\bm{x}_g\in\mathbb{R}^{n_x}$, as well as avoid the state set $\mathcal{X}_O\in\mathbb{R}^{n_x}$ occupied by obstacles:
\begin{equation} \label{eq:policy_safety_liveness}
    \begin{gathered}
        |\bm{x}(t_0+t_H)-\bm{x}_g| \le |\bm{x}(t_0)-\bm{x}_g| \\
        \bm{x}(t)\notin\mathcal{X}_O,\, \forall t'\in[t_0,t_0+t_H]
    \end{gathered}
\end{equation}
In addition to these requirements, this policy should optimize over a cost function $S(\tau)$ that may be nonlinear, non-smooth and even non-differentiable:
\begin{equation}
    \mathcal{P}_\tau =\argmin_{\tau}\left[ S(\tau) \right]
\end{equation}

Note that the commanded trajectory $\tau(t)$ may not be the same as the actual trajectory $\tau_\text{act}$ of the system over time, due to tracking error. To compensate for this, the proposed approach also considers a risk measure $\rho(\cdot)\in\mathbb{R}$ that relates a given trajectory $\tau(t)$ to the risk of collision due to this error $|\tau(t)-\tau_\text{act}(t)|$. Although we do not constrain $\rho$ to have any particular form, basic assumptions must be placed in order to cast the problem of risk minimization correctly: (i) $\rho(\cdot)$ is positive semi-definite, (ii) $\rho(\cdot)=0$ for situations that have no risk, and (iii) $\rho(\cdot)>0$ for situations that have risk of collision. With these assumptions, this risk can be included inside the cost function of the trajectory generation policy.

\textbf{Problem 2: \textit{Risk-aware Navigation}:}
Given a risk measure $\rho(\cdot)$, create a policy $\mathcal{P}_{\tau}$ for finding a trajectory $\tau$ that also minimizes risk:
\begin{equation}
    \mathcal{P}_\tau =\argmin_{\tau} \left[S(\tau) +  \int_\tau\rho(\cdot)dt\right]
\end{equation}
Next, we discuss our proposed approach
for safe, risk-aware navigation of a UAV by combining risk measures with a novel MPPI-based motion planning technique.


%% file: approach.tex
\section{Approach}\label{sec:approach}

\begin{figure}[ht!]
\centering
\includegraphics[width=0.47\textwidth]{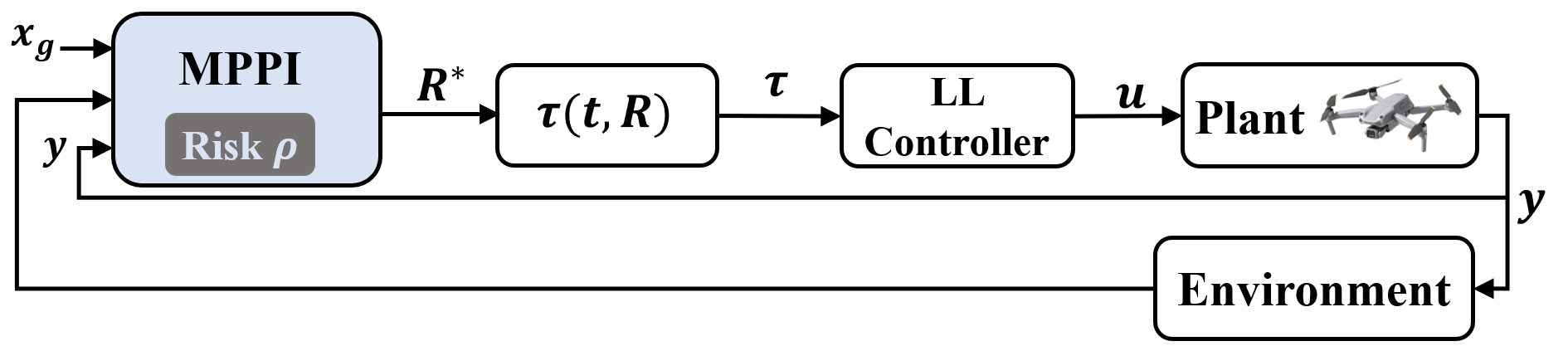}
\caption{Diagram showing the proposed motion planner (blue shaded cell) within the context of a general autonomy stack.}
\vspace{-5pt}
\label{fig:approach_diagram}
\end{figure}

We propose an MPPI-based motion planning policy that finds a trajectory $\tau(t)$ for the robot to track, which is chosen by the policy for its ability to guide the robot toward the goal state $\bm{x}_g$, while also minimizing the risk measure $\rho(\cdot)$. We consider trajectories defined by parameters $R$ so that the task of the MPPI planner is to find a set of parameters $R^*$ that optimize a control objective function over a future time horizon. The motion planner is sampled in a receding horizon fashion, allowing the robot to react to a potentially changing environment. Fig.~\ref{fig:approach_diagram} shows our approach within a typical autonomy stack, in which $\tau(t,R)$ is then fed to a low level controller that produces controls $\bm{u}$ for reliable tracking.

To facilitate discussion of the proposed motion planner, the scenario assumed by this paper involves a UAV traveling through a potentially cluttered obstacle course.
As is typical of the control pipelines for many AMR, there is some mismatch between the commanded trajectory $\tau(t)$ and the actual trajectory $\tau_\text{act}$, leading to some tracking error in the state space of the UAV. When the UAV is traveling in a cluttered environment, this mismatch could lead to collision with an obstacle. Thus, the MPPI motion planner should be aware of this risk and command trajectories that move towards the goal while remaining safe.

To this end, Sec.~\ref{sec:trajectory_parametrization} discusses the parameters $R$ that fully define a trajectory $\tau$. Sec.~\ref{sec:risk_measure} then discusses how $\rho(\cdot)$ can be concretely defined. Sec.~\ref{sec:mppi} discusses how an MPPI can utilize a general risk measure $\rho(\cdot)$ for fast, online and risk-aware motion planning. Finally, Sec.~\ref{sec:ensuring_safety_under_failure} discusses safety considerations under failure of the receding-horizon MPPI-based motion planner.
\subsection{Trajectory Parametrization} \label{sec:trajectory_parametrization}
The underlying MPPI motion planning algorithm relies on sampling different perturbations $\mathcal{E}$ to the parameters $R$, resulting in different trajectories defined by $\tau(t,R+\mathcal{E})$. These trajectories are continuously recomputed over some future horizon and applied in a receding-horizon fashion. In order to minimize the number of random samples needed for best results, the dimension of the sampling space $\mathcal{E}$ should also be relatively small.
This paper assumes $\tau$ takes the form of a minimum jerk trajectory, since it is a popular choice for UAV trajectory generation~\cite{7299672,8206119}. These minimum jerk trajectories are defined by $P$ waypoints $R=\{\bm{r}_1,\bm{r}_2,...,\bm{r}_P\}$, where $\bm{r}\in\mathbb{R}^3$ is the position of each waypoint, forming $P$ different trajectory segments.  Every min-jerk trajectory segment is defined by a fifth-order polynominal in time along each cartesian direction $i\in [x,y,z]$:
\begin{equation} \label{eq:min_jerk_polynomial}
    \tau^i(t) = \sum_{j=0}^5(c^i_j/j!)t^j
\end{equation}
Although the proposed approach may use any number of waypoints, the rest of the paper will restrict a single trajectory to be defined by $P=2$ waypoints. This is done to both simplify discussion and reduce the number of decision variables as much as possible, mitigating the computational burden of this approach.
This means a trajectory will have two distinct segments, where each is defined by a set of coefficients $\{c^i_{jk}\}$, with $j\in[0,5]$ denoting the associated power of $t$ and $k\in[1,2]$ denoting the trajectory segment. Fig.~\ref{fig:examples_of_trajectories}(a) shows an example trajectory defined by two waypoints, with each segment having an equal time span of $T$ seconds so that the total trajectory has a time span of $t_h = 2T$ seconds, equal to the horizon-length of the receding horizon planner.

The first trajectory segment (blue colored in Fig.~\ref{fig:examples_of_trajectories}(a)) is defined by the initial position $p^i_0$, velocity $v^i_0$ and acceleration $a^i_0$ of the UAV, as well as the position of the first waypoint $r^i_1$:
\begin{equation}
    \begin{gathered} \label{eq:traj_first_seg_bcs}
        p^i_0 = c^i_{01} \\
        v^i_0 = c^i_{11} \\
        a^i_0 = c^i_{21} \\
        r^i_1 = \sum_{j=0}^5 (c^i_{j1}/j!)T^j  \\
    \end{gathered}
\end{equation}
Likewise, the second trajectory segment (shown in orange in Fig.~\ref{fig:examples_of_trajectories}(a)) is defined by the first waypoint position $r^i_1$, the second waypoint position $r^i_2$, as well as the desired velocity $v^i_e$ and acceleration $a^i_e$ at the end of the total trajectory, as the UAV reaches the second waypoint. 
\begin{equation} \label{eq:traj_second_seg_bcs}
    \begin{gathered}
        r^i_1 = c^i_{02} \\
        r^i_2 = \sum_{j=0}^5(c^i_{j2}/j!)T^j \\
        v^i_e = \sum_{j=1}^5(c^i_{j2}/(j-1)!)T^{j-1} \\
        a^i_e = \sum_{j=2}^5(c^i_{j2}/(j-2)!)T^{j-2} \\
    \end{gathered}
\end{equation}
In order to avoid defining the velocity and acceleration conditions at the first waypoint, Pontryagin's minimum principle~\cite{7299672} may be used to allow these conditions to remain free. It can be shown that not constraining $v(t)$ and $a(t)$ at the first waypoint directly implies continuity to the associated costate variables $\lambda_{v}(t)$ and $\lambda_a(t)$, respectively. For a given direction $i$ and trajectory segment $k$, these costates are:
\begin{equation} 
    \lambda_{v}(t) = c^i_{4k}+c^i_{5k}t
\end{equation}
\begin{equation}
    \lambda_{a}(t) = c^i_{3k}+c^i_{4k}t+(c^i_{5k}/2)t^2
\end{equation}

Continuity of these costate values at the first waypoint amounts to the following constraints:
\begin{equation} \label{eq:traj_cont_costate}
    \begin{gathered}
        c^i_{41}+c^i_{51}T = c^i_{42} \\
        c^i_{31}+c^i_{41}T+(c^i_{51}/2)T^2 = c^i_{32}
    \end{gathered}
\end{equation}
Lastly, continuity of the velocity and acceleration at the first waypoint must also be enforced:
\begin{equation}\label{eq:traj_cont_vel_acc}
    \begin{gathered}
        \sum_{j=1}^5 (c^i_{j1}/(j-1)!)T^{j-1} = c^i_{12} \\ 
        \sum_{j=2}^5 (c^i_{j1}/(j-2)!)T^{j-2} = c^i_{22}
    \end{gathered}
\end{equation}

Together,~\eqref{eq:traj_first_seg_bcs},~\eqref{eq:traj_second_seg_bcs},~\eqref{eq:traj_cont_costate} and~\eqref{eq:traj_cont_vel_acc} form 12 linear equations of 12 unknown constants, which may be solved efficiently by any number of linear algebra solvers available~\cite{eigenweb}.

\subsection{Risk Measure} \label{sec:risk_measure}

Many types of risk measures have been used in different robotic applications, from defining $\rho(\cdot)$ as the probability of a fault for legged motion~\cite{9601255}, to conditional value at risk \cite{8767973}, a more sophisticated measure of risk that accounts for the severity of unlikely but possible events.

This paper proposes a data-informed risk measure that models geometric mismatch between the trajectory $\tau(t,R)$ tracked by the low level controller and the actual motion of the UAV, $\tau_\text{act}$.
Specifically, a relationship is established between this mismatch and the maximum speed commanded by $\tau(t,R)$. This relationship captures how higher robot speeds often  worsen tracking error of a desired trajectory by the low level controller. This degradation in performance can lead to unsafe situations, especially when the robot is travelling in a cluttered environment. Thus, the risk measure $\rho(\tau(t,R))$ relates the speed of the commanded trajectory to the risk of collision with nearby obstacles.

\begin{figure}[!ht]
\centering
 \subfigure[]{\includegraphics[width=0.218\textwidth]{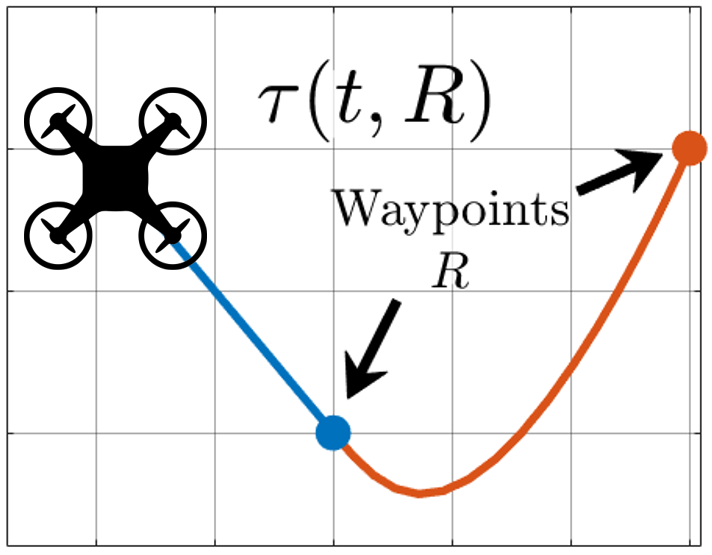}}\label{fig:minjerkEExample2}\hspace{1em}
 \subfigure[]{\includegraphics[width=0.22\textwidth]{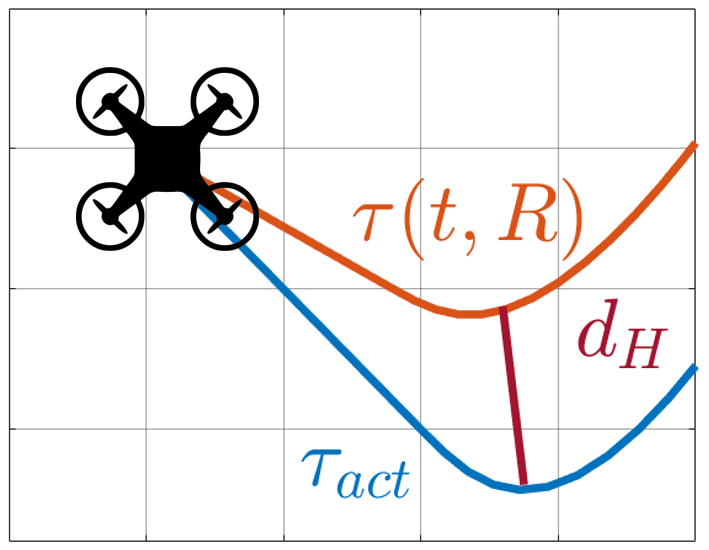}}\label{fig:hDistExample2}
   \vspace{-7pt}
\caption{Example of min-jerk trajectory shown in (a), and example of trajectory tracking error shown in (b).}
 \vspace{-5pt}
\label{fig:examples_of_trajectories}
\end{figure}

In order to define this risk measure, first define $d(t_1,t_2)$ as the euclidean distance between a point $\tau_1(t_1)$ on one trajectory and point $\tau_2(t_2)$ on another. The Hausdorff distance $d_H(\cdot)$ between two trajectories $\tau_1$ and $\tau_2$ is defined as:
\begin{multline}
    d_H(\tau_1,\tau_2) = \max\bigg\{ \max_{t_1\in[0,PT]}\left[\min_{t_2\in[0,PT]}d(t_1,t_2)\right],  \\ \max_{t_2\in[0,PT]}\left[\min_{t_1\in[0,PT]}d(t_2,t_1)\right] \bigg\}
\end{multline}
 Fig.~\ref{fig:examples_of_trajectories}(b) shows an example of how $d_H(\cdot)$ is found between the commanded trajectory $\tau(t,R)$ and the actual trajectory $\tau_\text{act}$ that results from trying to track $\tau(t,R)$.
 If $d_H(\tau(t,R),\tau_\text{act})\approx 0$, then both trajectories have considerable overlap in the $xyz$ space over the entire trajectory, while $d_H(\tau(t,R),\tau_\text{act})\gg 0$ signifies at least some portion
 where there is significant deviation. Through simulation or experiment, data can be collected that measures $d_H(\tau(t,R),\tau_\text{act})$ for various commanded trajectories.
 This data can then be used to train an estimate $\hat{d}_H\left(\tau(t,R)\right)$ that is only dependent on the commanded trajectory.
 For the specific UAV application considered in this paper we have observed - as intuitively expected - that the deviation is a function of the maximum speed $v_{max}$ commanded by $\tau(t,R)$. In this way it is possible to predict the tracking error using only information from the commanded trajectory.

 Denote $d_\text{obs}$ as the distance between $\tau(t,R)$ and the nearest obstacle. For the UAV, it is considered risky when $d_\text{obs} < \hat{d}_H$, since the deviation of the actual trajectory may extend toward the obstacle, potentially colliding with it. Likewise, if $d_\text{obs} \ge \hat{d}_H$, then there is no risk of collision, since the robot is expected to deviate from $\tau(t,R)$ by a distance smaller than the nearest obstacle.

 To this end, the risk measure is defined as:
 \begin{equation} \label{eq:risk}
     \rho(\tau(t,R)) = \max\left[ 0, \frac{\hat{d}_H}{d_\text{obs}} - 1\right]
 \end{equation}
 In this way, $\rho(\cdot)>0$ when there exists the potential for $\tau_\text{act}$ to intersect with the boundary of an obstacle, and $\rho(\cdot)=0$ when $d_\text{obs} \ge  \hat{d}_H$.

 

\subsection{MPPI For Motion Planning} \label{sec:mppi}
MPPI control is a sampling-based control method to find the solution to a stochastic optimal control problem (OCP). In the proposed approach, the MPPI solver is used to find a series of Cartesian waypoint positions $R=\{\bm{r}_1,\bm{r}_2\}$ that define a min-jerk trajectory $\tau(t,R)$.

The MPPI algorithm must be defined with respect to some stochasic equations of motion for the state $\bm{x}$:
\begin{equation} \label{eq:stochastic_dynamics}
    \bm{x}_{k+1} = \bm{f}(\bm{x}_k,\tau(t_k,R+\mathcal{E}))
\end{equation}
Here, $\bm{f}(\cdot)$ represents a discrete-time equation of motion in which the robot $\bm{x}$ evolves under the influence of a trajectory $\tau(t,R+\mathcal{E})$, where $\mathcal{E}=\{\bm{\epsilon}_1,\bm{\epsilon}_2\}$ are random perturbations on the Cartesian $xy$ position of the $p^\text{th}$ waypoint, with $\bm{\epsilon}_p\sim\mathcal{N}(0,\Sigma)$.
 

The stochastic optimal control problem may be defined as the minimization of an expectation value, denoted by $\mathbb{E}(\cdot)$:
\begin{equation} \label{eq:stochastic_ocp}
    R^* = \argmin_{R}\mathbb{E}_{\mathbb{Q}}\left[ S(R+\mathcal{E}) \right]
\end{equation}

The term $S(\cdot)$ defines the cost of the total trajectory, with waypoints $R$ being perturbed by stochastic variables $\mathcal{E}$ that create the probability distribution $\mathbb{Q}$. Sec.~\ref{sec:comps_of_cost} describes the components of this cost function, and Sec.~\ref{sec:solving_ocp} describes the MPPI algorithm for finding the solution to~\eqref{eq:stochastic_ocp}.

\subsubsection{Cost Function Components} \label{sec:comps_of_cost}

The total cost $S(\cdot)$ is defined over $P=2$ waypoints as:
\begin{equation}\label{eq:total_cost}
    S(R+\mathcal{E})=\phi(\bm{x}(PT)) + \int_0^{PT}\mathcal{C}(\bm{x}(t))dt
\end{equation}
The term $\phi(\cdot)$ is a terminal cost function defined as the error between the final state $\bm{x}(PT)$ and the goal state:
\begin{equation}\label{eq:terminal_cost}
    \phi(x(PT)) = w_g|\bm{x}(PT)-\bm{x}_g|
\end{equation}
The constant $w_g > 0$ is a scaling factor that is tuned to adjust the relative weight of the different objectives within~\eqref{eq:total_cost}.
The running cost $\mathcal{C}(\bm{x}(t))$ is defined by three terms:
\begin{equation} \label{eq:running_cost}
    \mathcal{C}(\bm{x}) = \mathcal{C}_{ct}(\bm{x}) + \mathcal{C}_{obs}(\bm{x}) + \mathcal{C}_{\rho}(\bm{x})
\end{equation}
The first term $\mathcal{C}_{ct}$ penalizes any violation of state and actuation constraints of the UAV. Because the UAV model is differentially flat, both the state and controls may be expressed as a function of $\bm{x}$ and its derivatives, meaning constraint violations can easily be checked and heavily penalized.
The second term $\mathcal{C}_{obs}$ is an obstacle cost that heavily penalizes collisions with known obstacles in the environment. Since MPPI is a gradient-free method, the exact form of $\mathcal{C}_{obs}$ can be quite sparse with gradient information, such as an indicator function used with an occupancy map. For the obstacle course assumed by this paper, the obstacle cost is defined using an indicator function $I_{o_j}(x)$ that returns $1$ when state $\bm{x}$ lies within obstacle $o_j$, else it returns zero.
\begin{equation}
    \mathcal{C}_{obs} = w_{obs}\sum_{j=0}^{n_o}\int_0^{PT}I_{o_j}(\bm{x}(t))dt
\end{equation}
The second term $\mathcal{C}_{\rho}$ is the portion of the cost related to the risk of a trajectory. Since we constrain $\rho(\cdot)$ to be positive semi-definite, the risk cost can be defined as proportional to this risk measure:
\begin{equation} \label{eq:risk_cost}
    \mathcal{C}_{\rho} = w_{\rho}\rho(\cdot)
\end{equation}
As was the case with the obstacle cost, using an MPPI-based approach to solving~\eqref{eq:stochastic_ocp} allows the exact form of $\rho(\cdot)$ to be quite flexible in its definition, since it does not require information about the gradient of $\rho(\cdot)$. For example, $\rho(\cdot)$ may be a  function that approximates some notion of risk that may be hard to write by hand, e.g., a learned policy trained on simulated or experimental data.

Inclusion of the risk measure defined by~\eqref{eq:risk} in the cost function has two different effects. First, trajectories are generally planned to be spatially away from obstacles in order to increase $d_\text{obs}$. Second, if the robot must move through small gaps in order to reach its goal, then trajectories that command slower motion are preferred in order to decrease $d_H$. These behaviors are more concretely explored in both simulation (Sec.~\ref{sec:simulation}) and experiment (Sec.~\ref{sec:experiment}) later in the paper.


\subsubsection{Solving the Stochastic OCP} \label{sec:solving_ocp}

\begin{figure}[t!]
\centering
\includegraphics[width=0.45\textwidth]{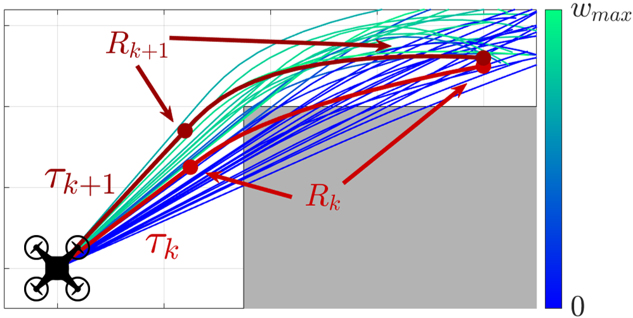}
\caption{An example iteration step of the MPPI algorithm.}
\vspace{-15pt}
\label{fig:mppiExample}
\end{figure}

Authors in~\cite{7989202} show how it is possible to obtain a theoretical \textit{exact} solution to~\eqref{eq:stochastic_ocp}. Unfortunately it is impossible to compute directly, but may be approximated using an iterative sampling method. This iterative algorithm relates the $(k+1)^\text{th}$ iteration to the $k^\text{th}$ iteration by:
\vspace{-3pt}
\begin{align}
    R_{k+1} &= R_{k} + \sum_{i=1}^{N}w(\mathcal{E}_i)\mathcal{E}_i \label{eq:weighted_sum}\\
    w(\mathcal{E}_i) &= \frac{1}{\eta} \exp\left[-S(R_k + \mathcal{E}_i)\right] \label{eq:weight_from_cost}
    \vspace{-3pt}
\end{align}
Here, $\eta$ is a normalization factor to ensure $\sum_i^{N}w(\mathcal{E}_i)=1$. Fig.~\ref{fig:mppiExample} illustrates how this algorithm finds a trajectory around an obstacle with $P=2$ waypoints. At the $k^\text{th}$ iteration, the current waypoint locations $R_{k}$ are subjected to a series of random perturbations $\{\mathcal{E}_i\}$. The blue/green trajectories are the results of these perturbations, with the color corresponding to the weight of the trajectory as determined by~\eqref{eq:weight_from_cost}. The $(k+1)^\text{th}$ iteration is found through a weighted sum of these perturbations, so that $\tau(t,R_{k+1})$ has a lower cost than the previous iteration. This procedure is repeated $n_\text{iter}$ times, and the resulting waypoints $R_{n_\text{iter}}$ are applied. In general, $n_\text{iter}$ is chosen to be as large as possible so that the iterative procedure can reasonably converge to the optimal solution.


\subsection{Ensuring Safety Under Failure} \label{sec:ensuring_safety_under_failure}

Lastly, it should be noted that due to the receding-horizon nature of the approach, a given trajectory may not be entirely traversed before another trajectory is replanned by the MPPI. Our approach leverages this by designing each trajectory segment shown in Fig.~\ref{fig:examples_of_trajectories}(a) with different purposes. The first trajectory segment (blue) commands basic acceleration, deceleration and coasting behaviors of the UAV, and the UAV is expected to track most if not all of this trajectory segment. However, the UAV is \textit{not expected to track the second trajectory segment (orange) under normal operation}. Instead, the MPPI should replan a new trajectory before the UAV reaches the first waypoint, thus ignoring the second trajectory segment altogether. If, however, the MPPI planner should fail, the UAV will continue to follow the second trajectory segment. In order to ensure the safety of the UAV, the second segment is designed so that the UAV safely stops at the second waypoint. This is easily done by setting the boundary velocity and acceleration of the second waypoint to zero, or $v^i_e=a^i_e=0$ in~\eqref{eq:traj_second_seg_bcs}.

%% file: simulation.tex
\section{Simulated Experiments}\label{sec:simulation}

Simulations were performed to validate the ability of the proposed approach to reduce risk and negotiate obstacles in a cluttered environment while navigating towards a goal. RotorS~\cite{Furrer2016} was used as a high-fidelity simulator for UAV motion that also includes a low level trajectory nonlinear controller~\cite{lee2010control}. Given a commanded trajectory $\tau(t)$, the low level controller attempts to track this trajectory, but due to measurement noise and physical limitations the actual trajectory $\tau_\text{act}$ is somewhat different, leading to a non-zero tracking error $d_{H}(\tau(t),\tau_\text{act}(t))$. In order to estimate $d_H(\cdot)$, data were collected by commanding different trajectories $\tau(t)$ and recording the resulting trajectory $\tau_\text{act}$. Each trajectory was defined by $P=2$ waypoints placed in the $xyz$ space, with $T=2.5$ seconds between each waypoints. Noise was injected into the simulated odometry sensor as a way to exacerbate the tracking error of the low level controller.

\begin{figure}[!ht]
\centering
 \subfigure[]{\includegraphics[width=0.22\textwidth]{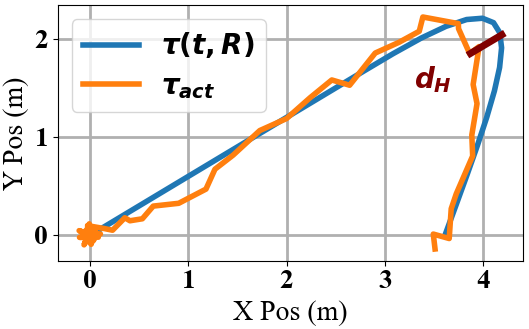}}\label{fig:setDiffExample}\hspace{1em}
 \subfigure[]{\includegraphics[width=0.22\textwidth]{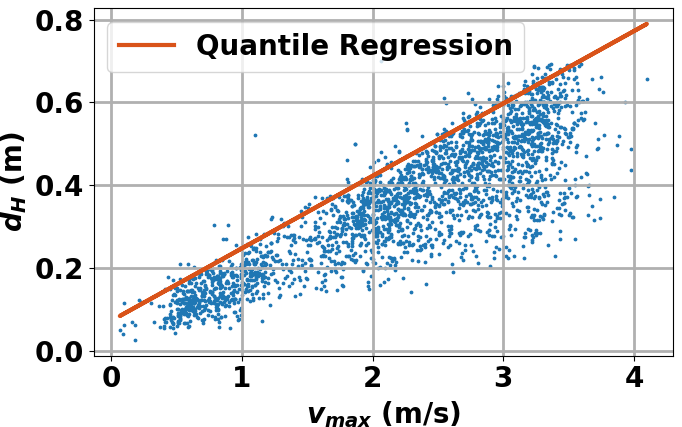}}\label{fig:setDiffLine}
   \vspace{-7pt}
\caption{RotorS UAV trajectory data used for risk cost $\mathcal{C}_{\rho}$.}
\label{fig:simSetDiff}
\end{figure}

Fig.~\ref{fig:simSetDiff}(a) shows a recorded example in which $\tau_\text{act}$ differs from $\tau(t)$. Also included in this plot is $d_{H}(\cdot)$ between the two trajectories. Fig.~\ref{fig:simSetDiff}(b) shows a plot of $d_{H}(\cdot)$ as a function of maximum speed $v_\text{max}$ along $\tau(t)$. It can be seen that as $v_{\max}$ increases, the set difference $d_{H}(\cdot)$ also increases.
A regression line $\hat{d}_{H}(v_{max})$ was fit to the data to capture the 95$^\text{th}$ percentile, giving a conservative estimate of the actual set difference, $\hat{d}_{H}(v_{max})\approx d_{H}(\cdot)$. This estimator of the set difference $\hat{d}_{H}$ was used in~\eqref{eq:risk} to find an estimate of the risk $\rho(\cdot)$.

\begin{figure*}[ht!] 
    \centering
    \subfigure[]{\includegraphics[width=0.222\textwidth]{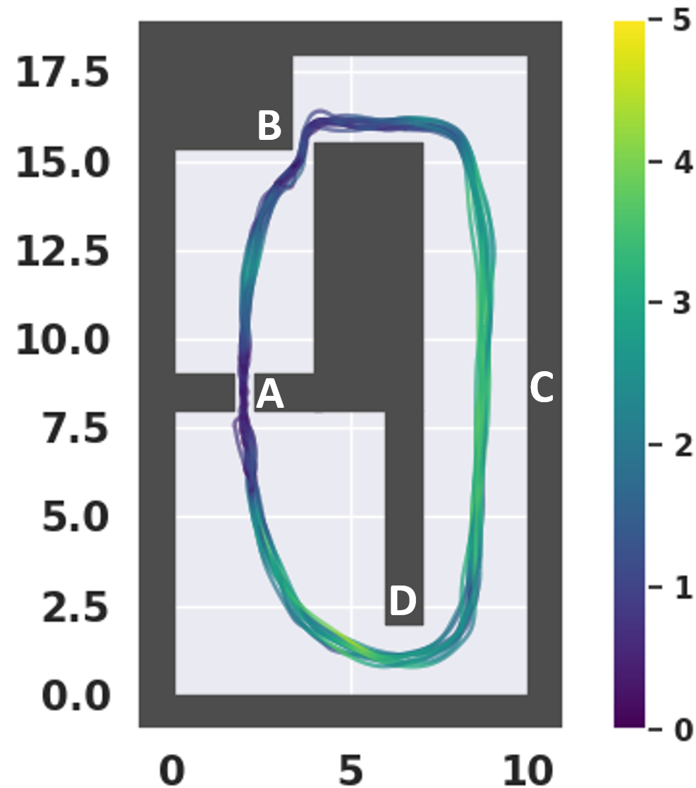}\label{fig:simWithRisk}} 
	\subfigure[]{\includegraphics[width=0.24\textwidth]{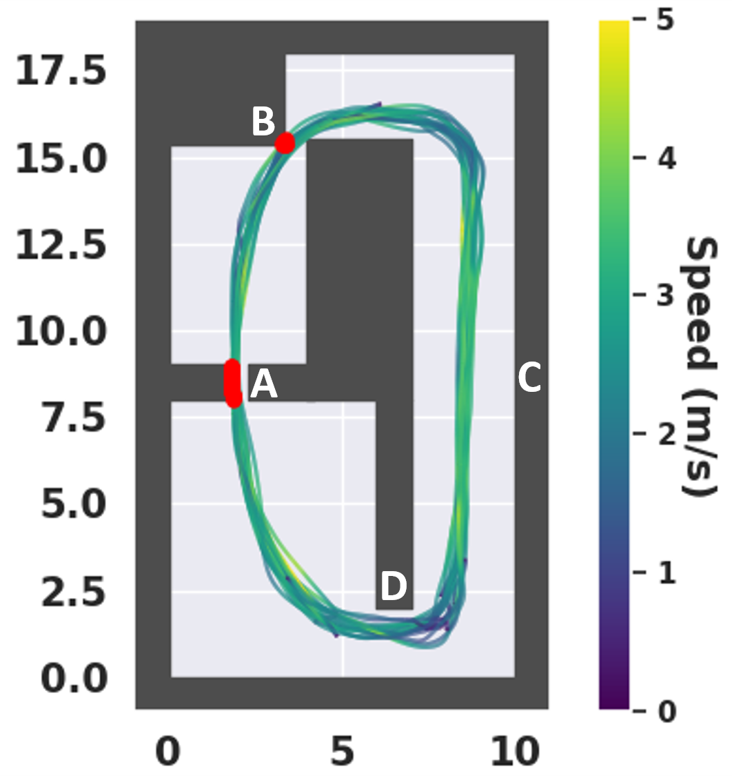}\label{fig:simWOutRisk}}\hspace{1em}
	\subfigure[]{\includegraphics[width=0.46\textwidth]{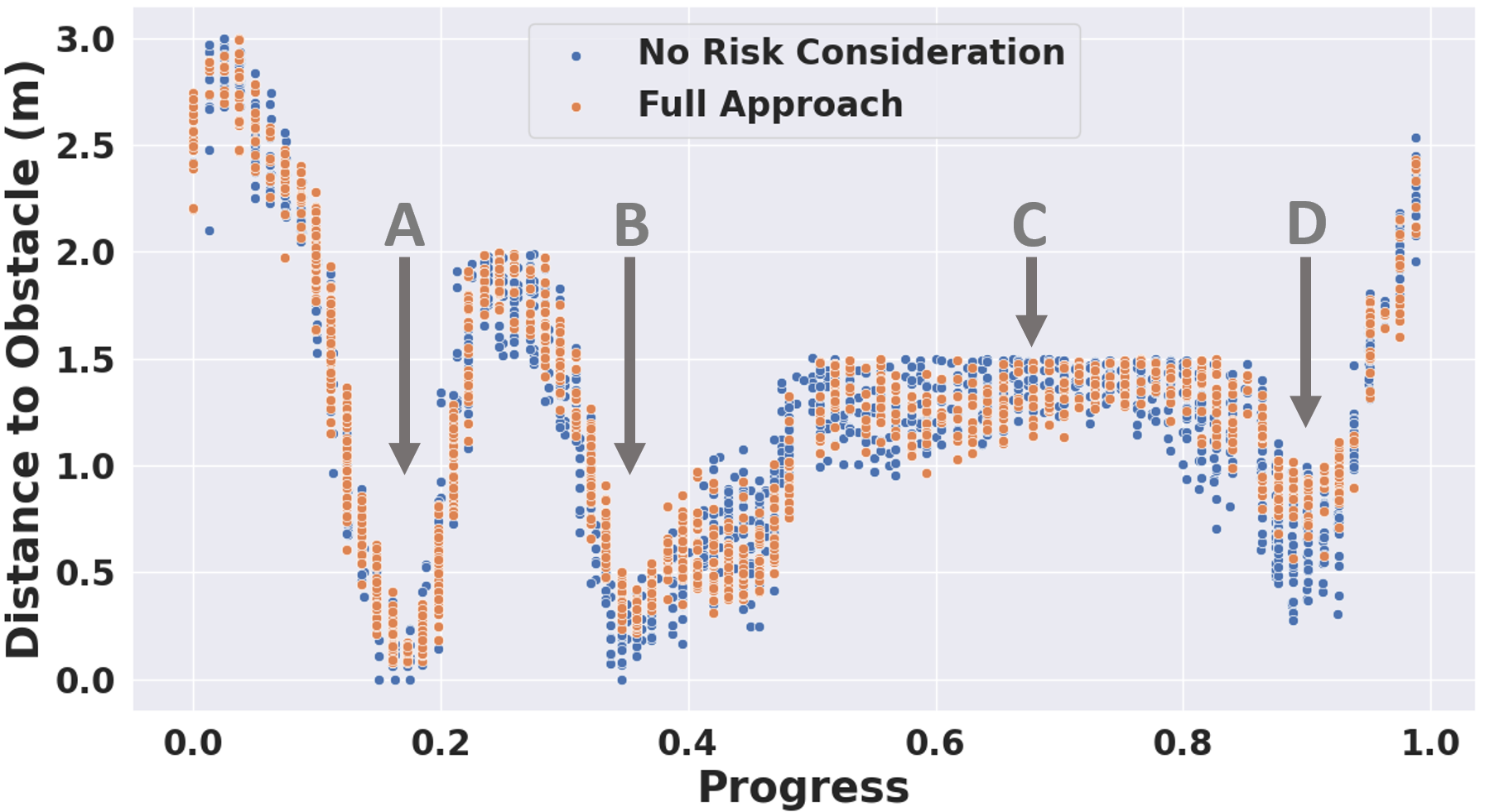}\label{fig:sim_dist_v_progress}}
 	\vspace{-8pt}
    \caption{Simulation of quadrotor navigating an obstacle course using the MPPI motion planner.}
     \vspace{-10pt}
	\label{fig:sim_results}
 	\vspace{-8pt}
\end{figure*}

The MPPI algorithm that solves~\eqref{eq:stochastic_ocp} was written in C++, using perturbation covariance matrix $\Sigma=diag(0.15,0.15,0.0)$~m, $N=50$ samples per iteration and $n_\text{iter}=200$ total iterations per MPPI sample.
The MPPI was run on a Lenovo ThinkPad X1 with Intel i7 6-core processor, and took an average of $0.19\pm0.03\text{s}$ to run. The algorithm took the current state of the UAV $\bm{x}_0$ as well as a set of local obstacles $\{\bm{o}_j\}$ and returned an optimal set of waypoints $R^{*}$ that defined a trajectory $\tau(t,R^{*})$ to be tracked. This trajectory was re-planned in a receding-horizon fashion, with the MPPI being re-sampled at a rate of $1\,\text{Hz}$ to find an updated set of waypoints.

To test the capabilities of the MPPI planner, the UAV was tasked with navigating an obstacle course, shown in Fig.~\ref{fig:sim_results}(a,b), with both small and large gaps to negotiate. To facilitate clockwise motion around the track, goal points were chosen a priori and commanded sequentially as the goal state in the MPPI cost function~\eqref{eq:terminal_cost}. These goal states also acted as a warm-start for the MPPI algorithm, choosing the initial waypoints $R_{0}$ to be along these goal points.


Fig.~\ref{fig:sim_results} also shows the result of 20 laps around the obstacle course. Fig.~\ref{fig:sim_results}(a) visualizes the resulting motion of the UAV as it tracked trajectories commanded by the MPPI. To highlight the effect of the risk objective on the overall motion of the UAV, Fig.~\ref{fig:sim_results}(b) shows the resulting trajectories with $w_{\rho}=0$ in the risk cost objective~\eqref{eq:risk_cost}. This effectively removed the consideration of risk within the MPPI when planning motion, instead finding a trajectory that avoided obstacles while moving as quickly as possible. Qualitatively, the difference between these trajectories is only apparent when the UAV approaches small gaps \textbf{A} and \textbf{B}. With our full risk-aware approach, the UAV slowed down enough to ensure safe passage through these tight spots, whereas the policy with no risk consideration sped through these gaps, resulting in collisions due to tracking error. These collisions are shown visually by the red dots in Fig.~\ref{fig:sim_results}(b). Alternatively, the risk-aware approach commanded the same high speed as the risk-agnostic policy through corridor \textbf{C}, since there were no close obstacles and it was safe to move quickly through this region.


Fig.~\ref{fig:sim_results}(c) additionally shows the distance between the UAV and the nearest obstacle as it traveled around the track. This distance is plotted against progress along the track, where $\text{progress}=0$ when the UAV was at the start of the course, and $\text{progress}=1$ at the end of the course.
This plot shows how the full approach worked to increase the distance between the UAV and obstacles, whereas the MPPI without risk consideration commanded motion closer to obstacles. Over the 20 laps, the full approach with risk consideration had 0 collisions, while motion with no consideration for risk resulted in 4 collisions.


%% file: experiment.tex
\section{Physical Experiments} \label{sec:experiment}

The proposed approach was verified experimentally on a Bitcraze Crazyflie quadrotor in two case studies: (i) a rectangular loop and (ii) a 4-way city block.
A Vicon motion capture system provided odometry information to an offboard laptop, which then used the MPPI path planner with the same parameters as in simulation to send trajectories to the Crazyflie's nonlinear controller \cite{Preiss2017Crazyswarm}. =Similar to the simulated experiments, trajectory tracking data were collected by commanding the Crazyflie to track various trajectories and recording $d_H(\tau(t,R),\tau_\text{act})$, which was used to train $\hat{d}_H(\tau(t,R))$. The effect of including the risk measure inside the MPPI cost function is shown by comparing the full approach to the same MPPI with $w_{\rho}=0$ in~\eqref{eq:risk_cost}.



\subsection{Rectangular Loop Case Study}

\begin{figure}[ht!]
\centering
\subfigure[No risk]{\includegraphics[width=0.22\textwidth]{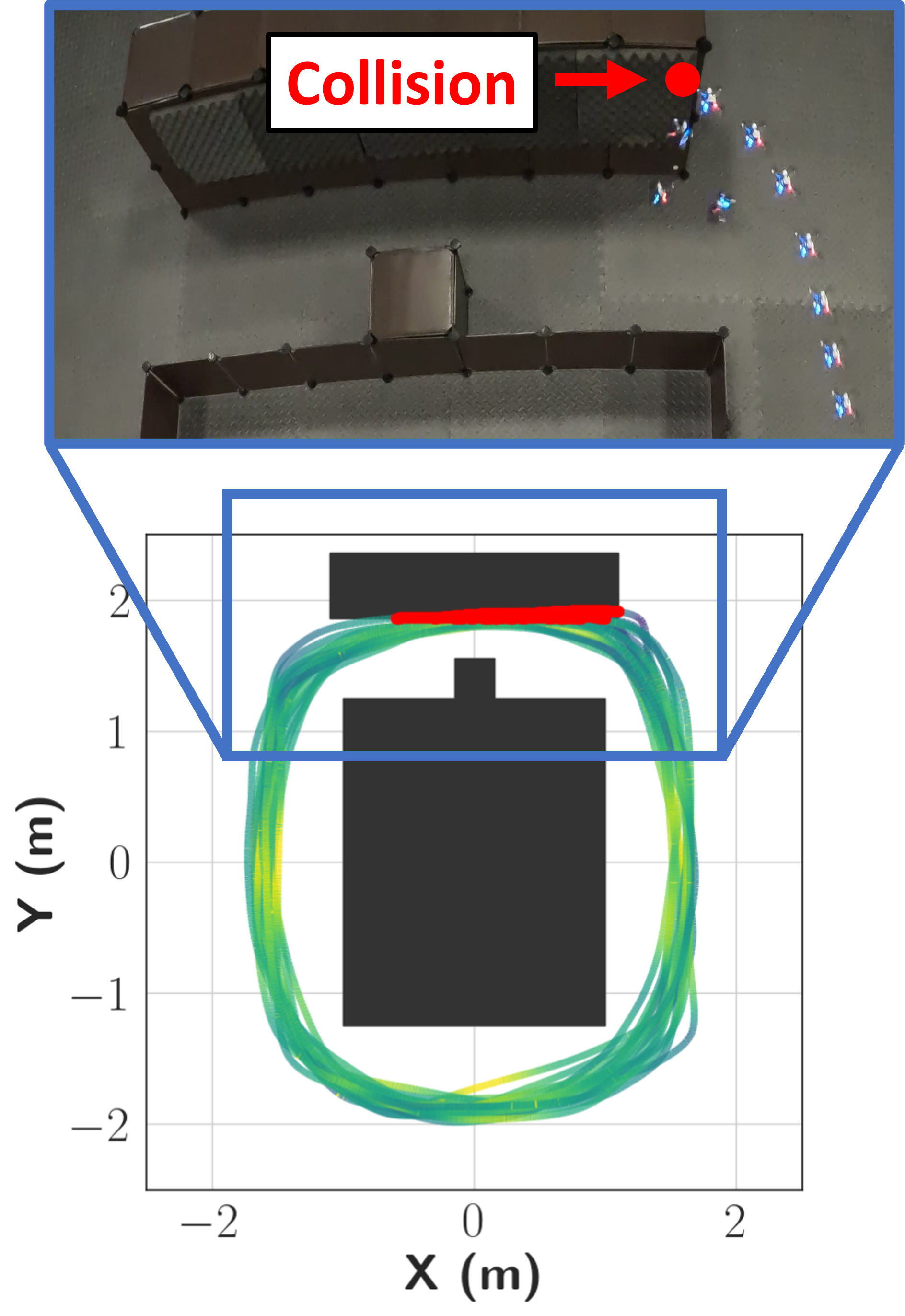}}\label{fig:crazyflieWithoutRiskExp2}
\subfigure[Full approach]{\includegraphics[width=.2345\textwidth]{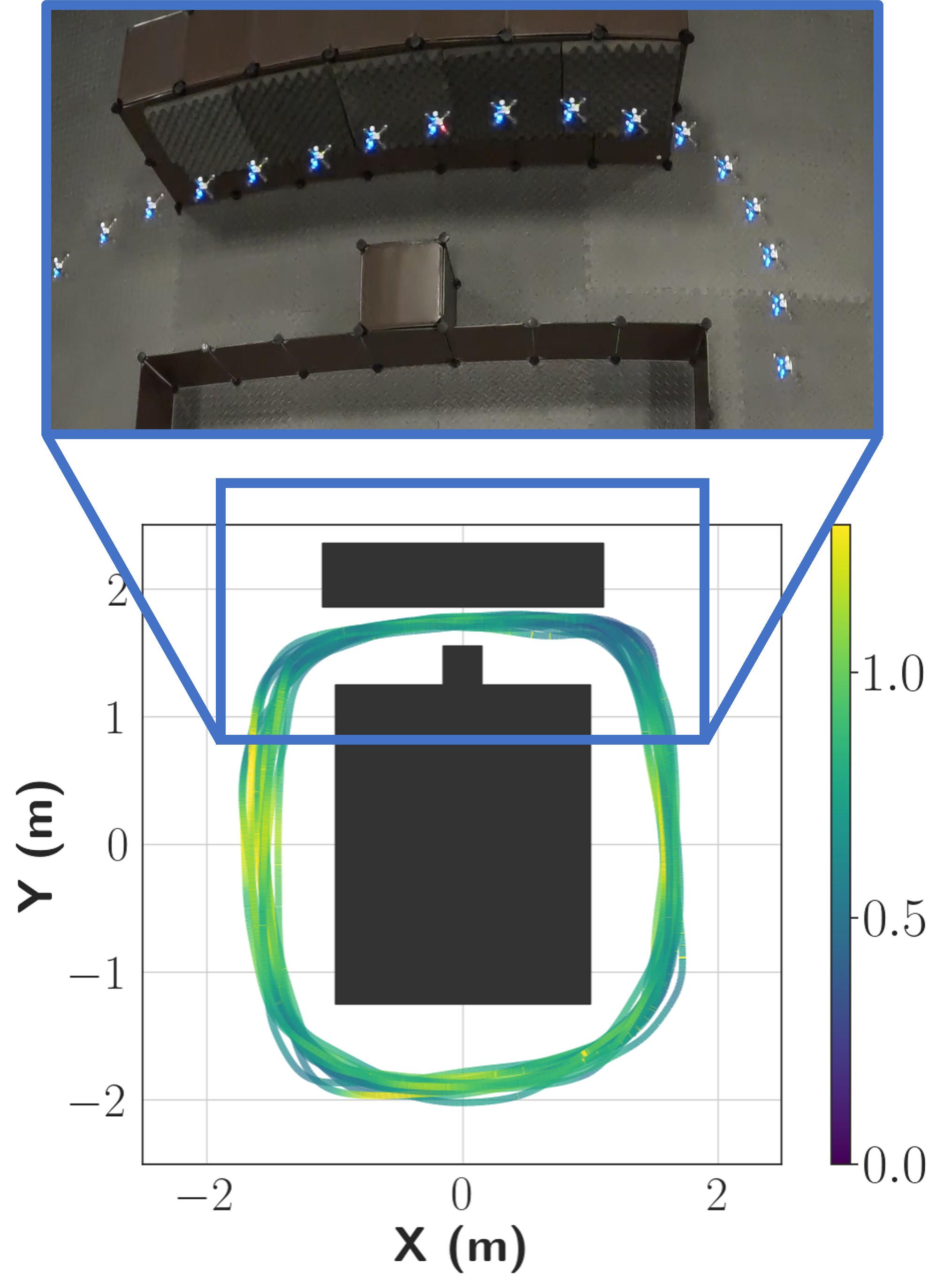}}\label{fig:crazyflieWithRiskExp2}\\[-1ex]

\caption{Crazyflie positions and velocities for the rectangular loop environment, along with snapshots of the physical experiment in our lab. }
\label{fig:experiment2Trajectories}
\vspace{-8pt}
\end{figure}

In this case study, the Crazyflie was tasked to complete loops around a central rectangular obstacle while negotiating its way through a narrow $30$~cm gap between the northern wall and a protruding square obstacle. The top portion of Fig.~\ref{fig:experiment2Trajectories} shows a snapshot of a sample pass through the narrow gap, both without risk consideration and with our full approach.  For these single trajectory examples, the physical obstacle height was raised to demonstrate a collision without risk consideration. For the rest of data collected, the physical obstacle height was lowered to allow multiple laps around the environment without disruption.
The results of the multiple laps are shown in the bottom plots of Fig.~\ref{fig:experiment2Trajectories}.
For this study, $20$ laps were recorded for both the no risk and full approach cases, giving $40$ total laps tested. Fig.~\ref{fig:experiment2Trajectories}(a) shows how the UAV collided with the north wall $6$ times (highlighted in red) when risk was not taken into account due to overshooting the planned MPPI trajectories at high speeds. However, with the full approach (Fig.~\ref{fig:experiment2Trajectories}(b)), no collisions occurred because the UAV slowed down at the bends of the loop in order to mitigate the overshooting behavior. Additionally, the UAV achieved comparable speeds both without risk consideration and with our full approach on the east, south and west side of the central square. This demonstrates how our full approach proactively adapted to move quickly through regions where there were no close obstacles, and the risk of collision due to tracking error was minimal.




\subsection{4-Way City Block Case Study}

\begin{figure*}[ht!]
    \centering
    \subfigure[]{\includegraphics[width=0.23\textwidth]{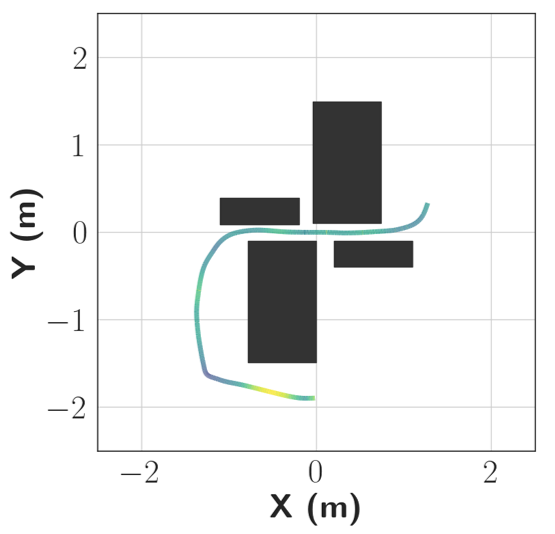}\label{fig:experimentPathA}} 
	\subfigure[]{\includegraphics[width=0.23\textwidth]{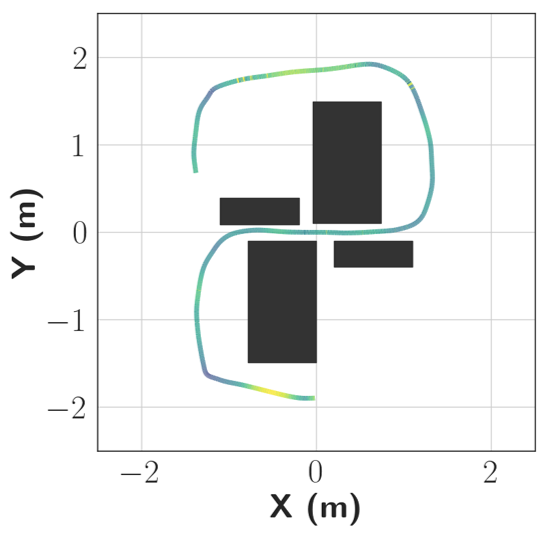}\label{fig:experimentPathB}}
	\subfigure[]{\includegraphics[width=0.23\textwidth]{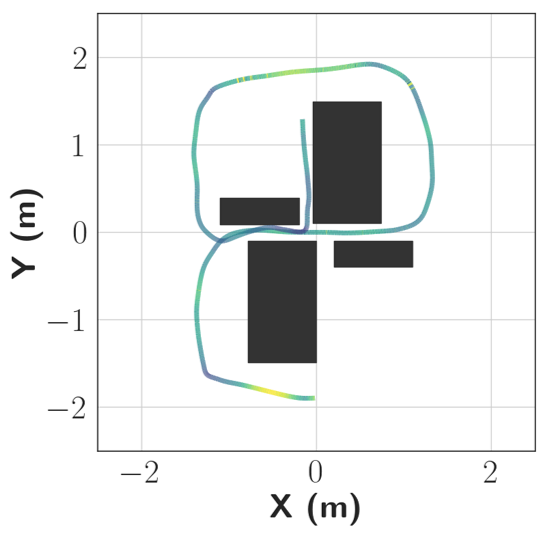}\label{fig:experimentPathC}}
	\subfigure[]{\includegraphics[width=0.268\textwidth]{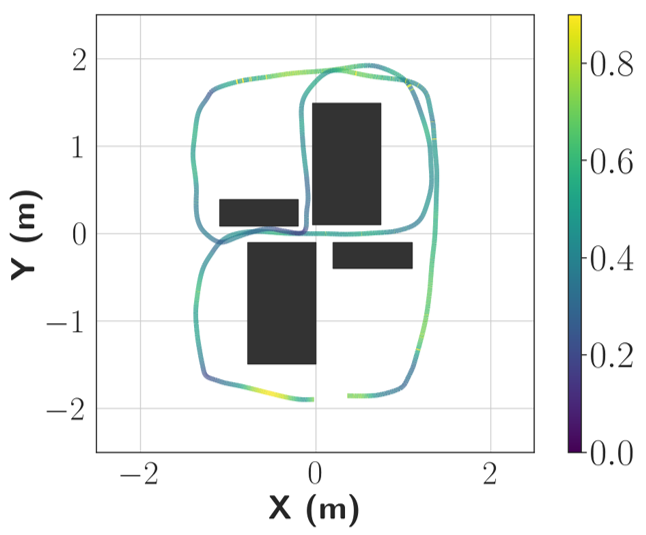}\label{fig:experimentPathD}}
 	\vspace{-6pt}
    \caption{A demonstration of a single lap that the UAV performs around the 4-way city block environment, along with its velocity profile.}
	\label{fig:crazyflieExperimentPath}
	\vspace{-12pt}
\end{figure*}

\begin{figure}[ht!]
\centering
\subfigure[No risk]{\includegraphics[width=0.22\textwidth]{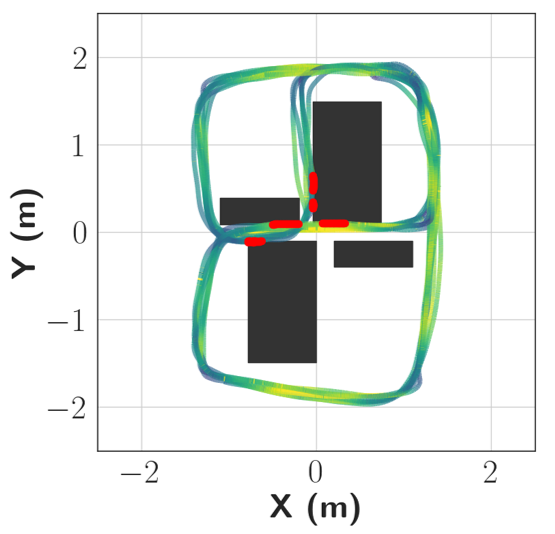}}\label{fig:crazyflieWithoutRisk}
\subfigure[Full approach]{\includegraphics[width=.254\textwidth]{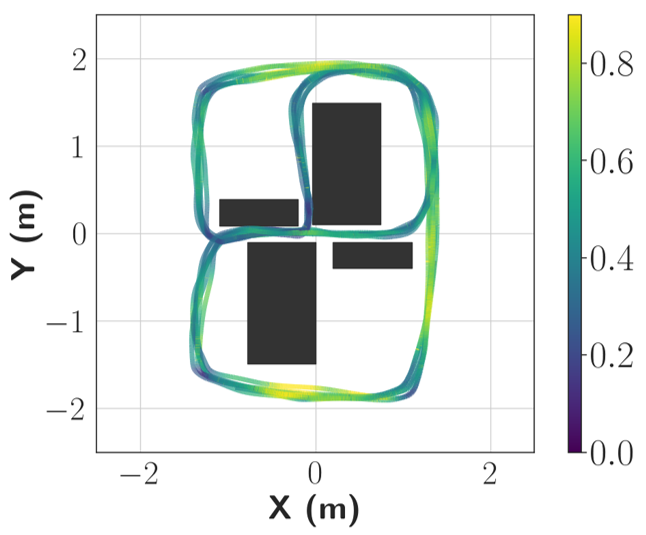}}\label{fig:crazyflieWithRisk}\\[-1ex]

\caption{Crazyflie positions and velocities for the 4-way city block environment (a) without considering risk and (b) with the full approach.}
\label{fig:experimentTrajectories}
\vspace{-10pt}
\end{figure}

In the second case study, the UAV was tasked to complete a complex path that involved alternating between straight lines and turning in different directions through a 4-way city block-like circuit. First, the UAV passed through a narrow $20$~cm horizontal channel in the center of the configuration (Fig.~\ref{fig:experimentPathA}). It then executed a u-turn around the narrow rectangular obstacle in the second quadrant (Fig.~\ref{fig:experimentPathB}) and went through a $20$~cm vertical passage (Fig.~\ref{fig:experimentPathC}) before looping back to the start (Fig.~\ref{fig:experimentPathD}). 
Fig.~\ref{fig:experimentTrajectories} shows the trajectory of the Crazyflie over $10$ laps in both the no risk and full approach cases, giving $20$ laps total. As can be seen from the results, in the no risk case (Fig.~\ref{fig:experimentTrajectories}(a)), the UAV collided with obstacles on $8$ occasions within the narrow corridors, while in the full approach (Fig.~\ref{fig:experimentTrajectories}(b)) it never collided. The speed profiles also demonstrate that, when obstacles are far enough away, the full approach allowed the UAV to reach the same speeds as with no risk consideration. Furthermore, thanks to our risk-aware framework, the UAV slowed down when traversing the cluttered sections of the environment, allowing for safer navigation that avoided collisions.

%% file: conclusion.tex
\section{Conclusions and Future Work} \label{sec:conclusion} 
In this work we have presented a receding-horizon path planning approach that can proactively adapt the trajectory of a robot at run-time in order to reduce overall risk while navigating through a cluttered environment. The proposed approach utilizes an MPPI control algorithm in order to accommodate a general, data-informed risk measure. Importantly, the trajectory planned by the MPPI is parameterized by a few number of variables, greatly reducing the computational requirements to run the MPPI algorithm and allowing the approach to run on more general hardware. The full approach was validated on a UAV robotic system navigating around obstacles towards a goal, with risk defined by the tracking error between commanded trajectory and the actual trajectory. Both simulation and experiment demonstrated how the inclusion of this risk measure inside the cost function allows the robot to move more safely through the environment, compared to motion without risk consideration.

Future work includes deploying this approach in additional robotic contexts, such as drones flying in outdoor environments with more complex sources of risk, or ground vehicles where risk may be associated with human-robot interaction. Additionally, the development of more sophisticated risk measures may facilitate more aggressive motion through cluttered environments, allowing the overall trajectory to be less conservative while ensuring risk is minimized.
\vspace{-1pt}

%% file: acknowledgement.tex
\vspace{-0pt}

\section{Acknowledgements}
\vspace{-2pt}
This work is based on research sponsored by Amazon Research Award and by DARPA under Contract No. FA8750-18-C-0090.
\vspace{-6pt}